%% file: 0_main.tex
\pdfoutput=1

\documentclass[11pt]{article}

\usepackage[preprint]{acl}

\usepackage{times}
\usepackage{latexsym}
\usepackage{booktabs}
\usepackage{graphicx}
\usepackage{booktabs}
\usepackage{amsmath}
\usepackage{amssymb}
\usepackage{array}
\usepackage{xcolor}
\usepackage{colortbl} 
\usepackage{float}
\usepackage{xspace}
\usepackage{multirow}
\usepackage[T1]{fontenc}

\usepackage[utf8]{inputenc}
\usepackage{enumitem} 
\usepackage{microtype}

\usepackage{inconsolata}

\usepackage{graphicx}

\newif\ifshowcomment
\showcommenttrue 

\newcommand{\vpara}[1]{\vspace{0.05in}\noindent \textbf{#1 }}

\newcommand{\model}{\texttt{PAS-SQL}\xspace}
\newcommand{\method}{PAS-SQL\xspace}

\usepackage{tcolorbox}
\tcbuselibrary{listings}
\tcbuselibrary{breakable}
\tcbuselibrary{skins}
\usepackage{multicol}
\usepackage{listings}
\usepackage{etoolbox}

\usepackage{fancyvrb}
\usepackage{framed}
\usepackage{verbatim}
\usepackage{fvextra}

\newcounter{myverbatimcounter}

\newtcblisting{myverbatim}[2][]{
  colback=gray!10,
  colframe=gray!50,
  boxrule=0.5pt,
  left=6pt,
  right=6pt,
  top=6pt,
  bottom=6pt,
  arc=0pt,
  boxsep=0pt,
  breakable,
  listing only,
  listing options={
    basicstyle=\ttfamily\small\linespread{3}, 
    keepspaces=true,
    columns=flexible,
    breaklines=true,
  },
  before upper*={\stepcounter{myverbatimcounter}%
    \par \vspace{2mm}\noindent%
    \begin{minipage}{\linewidth}
      \centering
      \textbf{Figure \themyverbatimcounter: #2}
    \end{minipage}%
    \vspace{2mm}
  },
  #1
}

\DeclareCaptionStyle{ruled}{labelfont=normalfont,labelsep=colon,strut=off} 
\floatstyle{ruled}
\newfloat{listing}{tb}{lst}{}
\floatname{listing}{Listing}

\title{Bridging the Gap: Transforming Natural Language Questions into SQL Queries via Abstract Query Pattern and Contextual Schema Markup}

\author{
 \textbf{Yonghui Kong\textsuperscript{1}},
 \textbf{Hongbing Hu\textsuperscript{1}},
 \textbf{Dan Zhang\textsuperscript{2}},
 \textbf{Siyuan Chai\textsuperscript{1}},
\\
 \textbf{Fan Zhang\textsuperscript{1}},
 \textbf{Wei Wang\textsuperscript{2}},
\\
\\
 \textsuperscript{1}Zhipu AI
 \quad
 \textsuperscript{2}Tsinghua University
}

\begin{document}
\maketitle

\begin{abstract}

Large language models have demonstrated excellent performance in many tasks, including Text-to-SQL, due to their powerful in-context learning capabilities. They are becoming the mainstream approach for Text-to-SQL.
However, these methods still have a significant gap compared to human performance, especially on complex questions. As the complexity of questions increases, the gap between questions and SQLs increases. We identify two important gaps: the structural mapping gap and the lexical mapping gap.
To tackle these two gaps, we propose \model, an efficient SQL generation pipeline based on LLMs, which alleviates gaps through Abstract Query Pattern (AQP) and Contextual Schema Markup (CSM).
AQP aims to obtain the structural pattern of the question by removing database-related information, which enables us to find structurally similar demonstrations.
CSM aims to associate database-related text span in the question with specific tables or columns in the database, which alleviates the lexical mapping gap. 
Experimental results on the Spider and BIRD datasets demonstrate the effectiveness of our proposed method.
Specifically, \model + GPT-4o sets a new state-of-the-art on the Spider benchmark with an execution accuracy of 87.9\%, and achieves leading results on the BIRD dataset with an execution accuracy of 64.67\%.
\end{abstract}

\input{1_introduction}
\input{3_method}
\input{4_experiments}
\input{2_related}
\input{5_conclusion}

\bibliography{coling}

\clearpage
\appendix

\input{6_appendix}

\end{document}

%% file: 1_introduction.tex
\section{Introduction}
With the widespread use of electronic devices, tables have become a common format for storing structured data from various sources, such as databases and spreadsheets~\citep{lu2024large}.
Natural language interfaces can help more non-technical users access the data while skilled professionals can efficiently access this data through Structured Query Language (SQL)~\citep{codd1974seven, deng2022recent,qin2022survey}.
Therefore, Text-to-SQL --- transforms natural language questions into executable SQL queries on databases --- has received extensive attention from both industry and academia~\citep{deng2022recent,qin2022survey,katsogiannis2023survey}.
Early methods primarily relied on rule-based systems that generate SQL queries through schema and template matching~\citep{androutsopoulos1995natural,zelle1996learning}, while these methods lack scalability and adaptability.
Recent methods aim to enhance domain independence by employing supervised models trained across different domains and datasets~\citep{scholak2021picard,qi2022rasat}. Nonetheless, this category of approaches suffers from poor generalization capacity and necessitates retraining when adapted to various databases. 

\begin{figure}[tbp]
    \centering
    \includegraphics[width=0.9\linewidth]{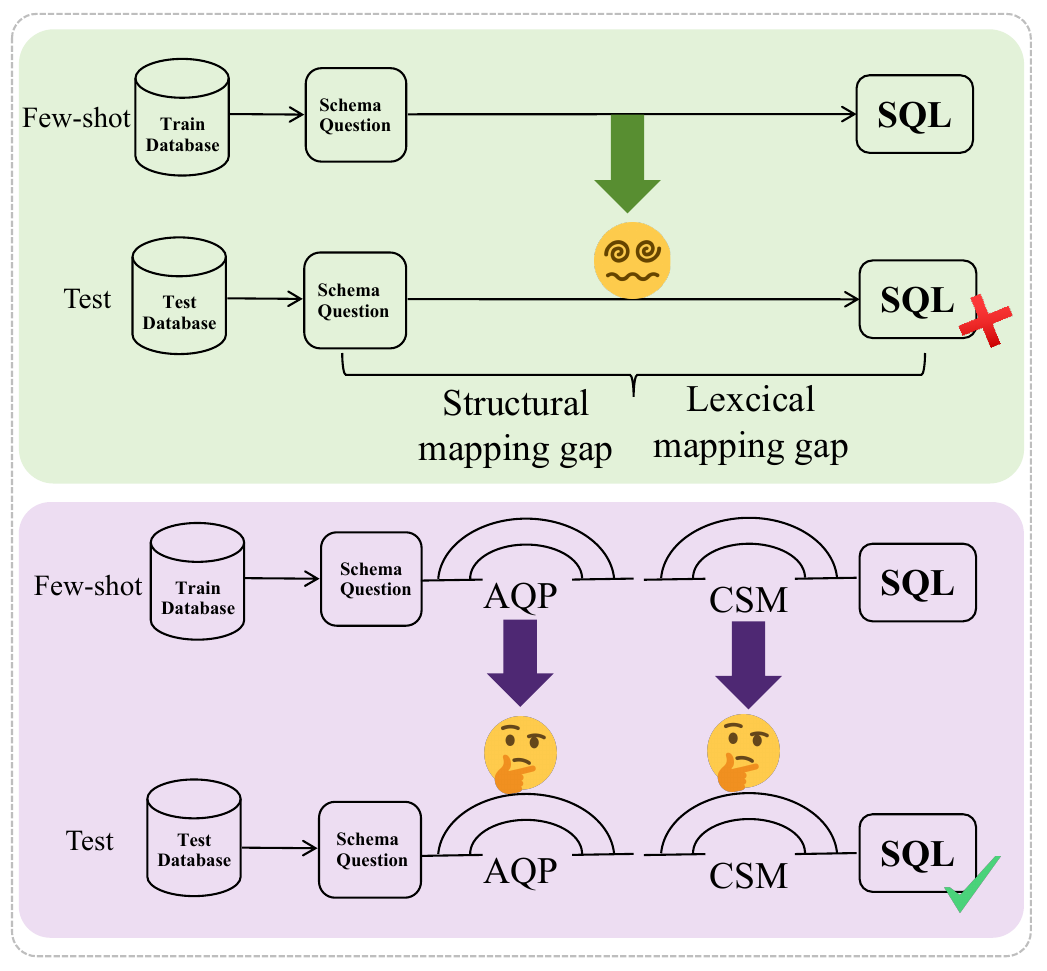}
    \caption{\model alleviates gaps through AQP and CSM. AQP and CSM denote Abstract Query Pattern and Contextual Schema Markup respectively.}
    \label{fig:enter-label}
    \vspace{-6mm}
\end{figure}

Recently, the advent of LLMs has led to significant advancements in Text-to-SQL task~\citep{rajkumar2022evaluating,ni2023lever,gao2024text}. 
Methods based on LLMs can be roughly classified into fine-tuning based methods and prompt-based methods.
Fine-tuning based methods mainly involve further training open-source language models using Text-to-SQL data~\citep{li2024codes,pourreza2024dts}. 
Prompt-based methods utilize in-context learning ability of closed-source language models to accomplish text-to-SQL tasks~\citep{pourreza2024din,gao2024text,li2024dawn,qu2024before}.
Fine-tuning-based methods usually require a certain amount of labeled data and computational resources.
With the continuous improvement of closed-source models, even after training on large amounts of data, some fine-tuning based methods on open-source language models are still weaker than prompt-based methods. Therefore, we focus on prompt-based methods.

Although prompt-based methods have achieved excellent performance on Text-to-SQL, their performance still has a significant gap compared to human performance, especially on complex questions.
We analyze and find that as the complexity of questions increases, the complexity of the corresponding SQLs also increases, and the gap between questions and SQLs becomes larger and larger.
We regard Text-to-SQL as a natural language to SQL translation task and identify two gaps from the perspective of translation: \textbf{(1) Structural mapping gap}. 
By removing the database-related information from the question and SQL, we can obtain the structural information of the question and the SQL. 
To generate SQL, the model needs to learn the mapping from the question structure to the SQL structure. As the complexity of the question increases, the structural mapping gap widens, making it more difficult for the model to learn.
\textbf{(2) Lexical mapping gap}. The more complex the question, the more database tables and columns are involved in the question, making it significantly more difficult to correctly map database-related text to the database schemas.


To tackle these two gaps, we propose \model, a Text-to-SQL system designed for complex and real-world databases. \model introduces a scalable and efficient SQL generation pipeline based on LLMs, consisting of four components: \textit{Abstract Query Pattern (AQP), Contextual Schema Markup (CSM), Constructing Demonstrations, Generating and Correcting SQL}.
AQP aims to obtain the structural pattern of the question by removing database-related information.
Specifically, we mask database-related information in the question using placeholders (e.g., [TABLE], [COLUMN], and [VALUE]) to obtain AQP representation.
The AQP representation of the question provides a database-agnostic representation that focuses on structure, which enables us to find structurally similar demonstrations, independent of the specific database. 
CSM aims to associate database-related text span in the question with specific tables or columns in the database to obtain CSM representation.
The CSM representation provides an effective method for integrating relevant database schema in the question, which alleviates the lexical mapping gap.
We process question-SQL pairs to obtain AQP and CSM demonstrations through the AQP and CSM modules.

To utilize AQP and CSM demonstrations to alleviate the two gaps, the generation process is divided into three steps: first, generate the AQP; next, generate the CSM; and finally, generate the SQL based on the AQP and CSM.
Specifically, when generating the CSM representation of a test question, we retrieve structurally similar demonstrations as few-shot examples based on the AQP representation to assist in CSM generation.
Inspired by the Chain of Thought (CoT) approach and aware of the high token consumption, we propose the CoT version, which generates AQP, then CSM, and finally SQL in a step-by-step way. 
We conduct extensive experiments on two datasets and our proposed method achieves an accuracy of 64.67\% on the BIRD dev set and 87.9\% on the Spider dev set.

Our key contributions can be summarized as follows:
\begin{itemize}[leftmargin=*,itemsep=0pt,parsep=0.5em,topsep=0.3em,partopsep=0.3em] 
\item We identify two important gaps when generating SQL queries for complex questions: the structural mapping gap and the lexical mapping gap.
\item We propose an efficient SQL generation pipeline, which effectively alleviates the two gaps through Abstract Query Pattern and Contextual Schema Markup module.
\item \model achieves impressive results, with an execution accuracy of 64.67\% on the BIRD dev set and 87.9\% on the Spider dev set.
\end{itemize}

%% file: 3_method.tex
\section{Method}

\begin{figure*}[htbp]
    \centering
    \includegraphics[width=0.95\textwidth]{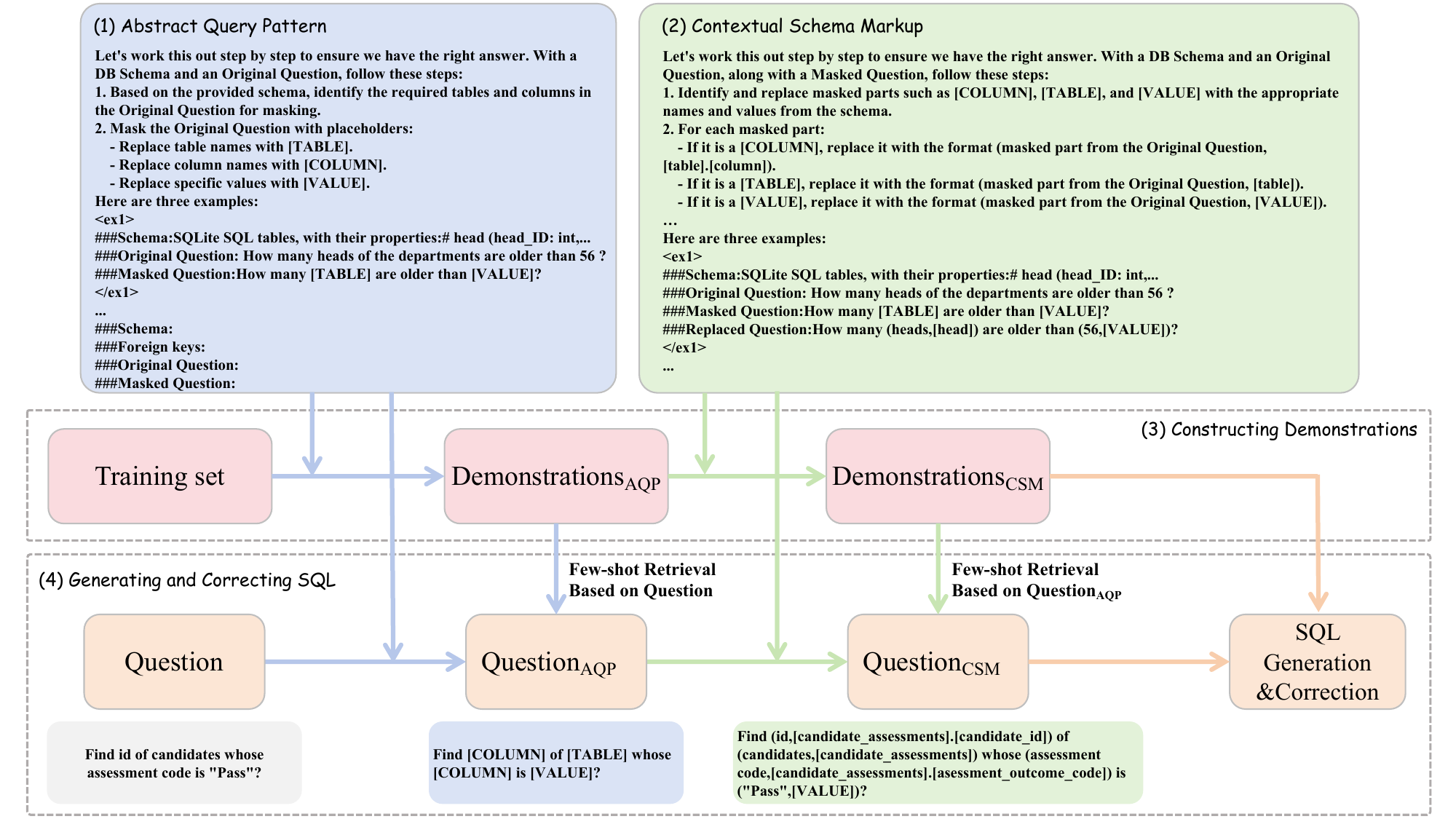}
    \caption{Overall framework of proposed \model.}
    \label{fig:method}
    \vspace{-4mm}
\end{figure*}

As the complexity of questions increases, the complexity of the corresponding SQLs also increases, and the gap between questions and SQLs becomes larger and larger. To bridge the gap, we propose \model as shown in Figure~\ref{fig:method}, which consists of four components: Abstract Query Pattern, Contextual Schema Markup, Constructing Demonstrations, Generating and Correcting SQL.
Additionally, we introduce a Chain-of-Thought (CoT) version of \model to reduce token usage.

\subsection{Abstract Query Pattern}
\label{method_AQP}

Specifically, there are two gaps: the mapping from natural language grammar structure to SQL grammar structure, and the mapping from natural language text to database schemas. 
To alleviate the first gap, we propose an Abstract Query Pattern (AQP) module. As illustrated in Figure~\ref{fig:method}, AQP aims to obtain the structural pattern of the question by removing database-related information. Next, we introduce how to get AQP representations of natural language questions. 

We conduct AQP by prompting the LLM. 
First, we give task instructions. Specifically, the model needs to identify the database-related information within the question and then replace it with three placeholders: \textit{[TABLE], [COLUMN], and [VALUE]}. 
Second, we provide pre-processed demonstrations that already include the AQP representations of questions to help the model better conduct AQP tasks. 
Finally, we concatenate task instructions, few-shot demonstrations, database schemas, and the question to obtain the AQP prompt.
The LLM process the AQP prompt to generate the AQP representation for the question.

The AQP representation of the question provides a database-agnostic representation that focuses on structure.
We claim that questions with similar AQP representations have similar SQL structures and we confirm this in the experiment. 
Therefore, we can retrieve structurally similar few-shot examples based on AQP, which can effectively help generate SQL for the test question.


\subsection{Contextual Schema Markup}\label{method_CSM}

To alleviate the second gap, we propose a Contextual Schema Markup (CSM) module. As illustrated in Figure~\ref{fig:method}, CSM aims to associate database-related text span in the question with specific tables or columns in the database. 
For the second gap, previous research has primarily concentrated on schema-linking, extracting necessary database schemas for SQL generation, thereby reducing the impact of a large number of unrelated schemas.
However, most of these works perform schema-linking at the question level, while we perform schema-linking at the span level. 

We conduct AQP by prompting the LLM. 
First, we give task instructions. The model needs to determine the database schema corresponding to each placeholder in the AQP representation of the question and replace the placeholder with the original text and the corresponding database schema.
Second, we provide pre-processed demonstrations that already include the CSM representation of questions to help the model better conduct CSM tasks.
Finally, we concatenate database schemas, the question with task instructions, and few-shot examples to obtain the CSM prompt. The LLM then processes this prompt to generate the CSM representation for the question.

By matching database-related text span with the tables and columns in the schema, CSM provides an effective method for integrating relevant schema in the question. 
We believe the CSM module can provide the necessary database schemas to alleviate the semantic gap, while AQP can provide relevant structures to alleviate the structural gap. With the combined effect of the two modules, SQL generation for complex questions can be better addressed.

\subsection{Constructing Demonstrations}

Both the Spider and BIRD datasets have a large number of training samples, and existing works directly use training samples as demonstrations. 
We claim that this approach does not make full use of these valuable natural language questions and SQL pairs.
We construct AQP and CSM demonstrations from the training samples to better utilize them.

We first select five representative samples from each training set and manually annotate them to obtain their AQP and CSM representations. Thus we get five AQP demonstrations and five CSM demonstrations for each dataset.
We combine database schemas, the question to be processed, SQL, and AQP demonstrations to obtain the AQP prompt, which is processed by LLM to obtain the AQP representation as described in Section~\ref{method_AQP}.
Next, we use the AQP representation of the question and CSM demonstrations to generate the CSM representation as described in Section~\ref{method_CSM}.
In these two steps, we use GLM-4 because the SQL corresponding to the question is already in the prompt, both tasks are relatively simple.

\subsection{Generating and Correcting SQL}

To generate the final SQL for the test question, we follow four steps. First, we obtain the AQP representation of the question. Next, we derive the CSM representation. Then, we generate the initial SQL based on the AQP and CSM representations. Finally, we validate and correct it to obtain the final SQL. Below, we introduce each step in detail.


\vpara{Generating AQP representation.} 
For the given test question, we start by retrieving the top-k most similar AQP demonstrations from all available AQP demonstrations as few-shot examples, using question similarity as a criterion for selection. Existing works indicate that the order of examples significantly affects the output. Therefore, we arrange the examples in reverse order, placing the most similar example closest to the test question.
By concatenating the task instructions, few-shot examples, database schemas, and the test question, and then processing them through an LLM, we obtain the AQP representation of the question as described in~\ref{method_AQP}. 

\vpara{Generating CSM representation.} Given the AQP representation of the question, we retrieve the most similar CSM demonstrations as few-shot examples based on AQP representation similarity. 
We then integrate the task instructions, few-shot examples, database schemas, and the test question to construct the CSM prompt. Subsequently, we generate the CSM representation of the question as outlined in Section~\ref{method_CSM}.

\vpara{Generating SQL.} 
We obtain the AQP representation and CSM representation of the question through the above two steps.
We then retrieve values from the database, ensuring that the relevant columns are present and emphasized in the CSM representation. 
We select three demonstrations as few-shot examples from the results of the previous CSM step. 
We combine the task instructions, few-shot examples, database schemas, the question, the AQP representation, and the CSM representation to obtain the SQL generation prompt.
This prompt is fed into the LLM to generate the SQL.

\vpara{Correcting SQL.} 
We further validate the generated SQL to ensure its executability.
We execute the generated SQL, and if it executes successfully, the generation process ends. Otherwise, we initiate the correction process.
The initial LLMs input, incorrect SQL, and detected errors are re-input into the LLMs to obtain a new SQL. 
This process repeats until the generated SQL can be successfully executed or a maximum number of correction attempts is reached.

\subsection{Chain-of-Thought}
Inspired by the Chain of Thought (CoT) approach and aware of the high token consumption from multiple calls of the language model, we propose the CoT version of \model. 
We first retrieve similar examples using the test question and then integrate the examples with the question to guide the model in sequentially generating AQP, CSM, and SQL. The results show that this approach significantly reduces token usage while maintaining minimal performance loss.
Previous research has applied Chain-of-Thought to the Text-to-SQL domain by decomposing problems into multiple steps for step-by-step resolution. 
However, as far as we know, there is no method to apply a structured chain of thought structure like ours.

%% file: 4_experiments.tex
\section{Experiment}

\subsection{Experimental Setup}

\subsubsection{Evaluation Datasets}

\vpara{Spider.} Spider is a cross-domain Text-to-SQL benchmark including 10,181 natural language questions and 5,693 unique SQL queries across more than 200 databases in 138 different domains~\citep{yu2018spider}. The dataset is split into the training set (7,000 examples), the development set (1,034 examples), and the test set (2,147 examples).

\vpara{BIRD.} BIRD is a large-scale cross-domain benchmark~\citep{li2024can}, which contains over 12,751 unique question-SQL pairs, and 95 large databases, covering 37 domains with a total size of 33.4GB. 
Compared to Spider, BIRD emphasizes SQL efficiency and knowledge reasoning in large databases, offering configurations with and without external knowledge. 
The training set has 9,428 examples and the development set has 1,534 examples.

\subsubsection{Metrics}

We evaluate model performance using the official metrics for each dataset.
Spider uses Execution Accuracy (\textbf{EX}), which compares the execution results of the predicted SQL with the actual results, offering a more precise performance measure.
BIRD uses Valid Efficiency Score (\textbf{VES}) and EX. VES evaluates both execution accuracy and run-time efficiency of the generated SQL, ensuring the results match the reference query and accounting for execution time efficiency.
We compare our approach with existing works, including DIN-SQL~\citep{pourreza2024din}, DAIL-SQL~\citep{gao2024text}, MAC-SQL~\citep{wang2024mac},TA-SQL~\citep{qu2024before}, SuperSQL~\citep{li2024dawn}. 

\subsubsection{Implement Details}

In our experiments, we use GLM-4~\citep{glm2024chatglm} to process the training sets to obtain AQP and CSM demonstrations. 
In the prompt, we use three few-shot examples for both datasets because our experiments indicate that using three yields better results, as depicted in Figure \ref{fig:fewshot_number}.
For retrieving similar few-shot examples, we utilize the BGE model~\citep{zhang2023retrieve} and conduct embedding similarity search with the FAISS library~\citep{douze2024faiss}.
To minimize the randomness in the outputs of the large language models (LLMs), we set the temperature to 0. 

\input{tables/bird_dev}

\subsection{Overall Performance}

\subsubsection{BIRD Results}

As shown in Table \ref{tab:bird_dev_results}, on the BIRD Dev set, our method \method+ GPT-4o achieves a significant improvement, with an execution accuracy of 64.67\%, nearly 10\% higher than DAIL-SQL+ GPT-4. 
\method+ GPT-4o also demonstrates a notable enhancement over the GPT-4o baseline, reaching a new SOTA. 
For the Valid Efficiency Score (VES) metric, \method+ GPT-4o achieves 65.04\%, surpassing other methods, indicating the high efficiency of SQL generated by our method.
These results indicate that our method can find more valuable few-shot examples for test questions by abstracting training samples. This approach allows for more effective utilization of the knowledge contained within the training data, leading to the generation of more accurate and efficient SQL queries.
In addition, while GLM-4 is only 2.47\% behind GPT-4o, the introduction of \method increases this gap to 5.54\%. 
Although we obtain AQP and CSM few-shot examples using GLM-4 and apply these few-shot examples to GPT-4o, our method significantly enhances GPT-4o's performance. 
These performance improvements across different language models demonstrate the model-agnostic nature and strong adaptability of \method.

\input{tables/spider_results}

\subsubsection{Spider Results}

As shown in Figure~\ref{tab:spider_results}, our method significantly outperforms other prompt-based methods.
On the Spider Dev set, our method achieves an execution accuracy of 87.9\%, outperforming GPT-4o by 12.9\% and the current state-of-the-art DAIL-SQL + GPT-4 by 3.5\%, setting a new SOTA. We also achieve SOTA on the Spider Test set. 
Additionally, \method+ GLM-4 delivers competitive results.
These results further demonstrate our method's effectiveness and generalization. 
Interestingly, we observe that the performance of GLM-4 is very close to GPT-4o on Spider, even slightly outperforming GPT-4o. 

\input{tables/bird_different_level}

\subsection{Ablation Study}
\subsubsection{Major Components}

We conduct ablation studies to assess the impact of the AQP and CSM components of our proposed method. The results are illustrated in Figure~\ref{tab:bird_different_level}. 
Removing the CSM component leads to a 2.94\% drop in performance, highlighting that the absence of question and database schema mapping makes it more challenging for the model to generate SQL, thus lowering accuracy.
CSM is equivalent to performing schema-linking based on few-shot examples, which can effectively reduce the number of candidate database schemas in the next step.
Without CSM, when generating SQL, the model needs to handle a large number of redundant and unrelated database schemas, making it more difficult.

Removing AQP along with CSM causes a further performance drop of 4.57\%, indicating the importance of AQP. 
AQP not only facilitates the generation of CSM but also plays a key role in retrieving similar few-shot examples. 
By masking database-related information in the original question, AQP standardizes the question, making it easier for the model to find structurally similar few-shot examples from a vast array of training samples. 
Without AQP, the model struggles to obtain relevant few-shot examples, leading to a marked drop in SQL execution accuracy.

\input{tables/bird_ablation}

\subsubsection{Minor Components}
We conduct ablation studies on several specific aspects of \method. Initially, we investigate the impact of the organization of few-shot examples on performance. In our approach, the few-shot examples retain only the tables and columns used in the example SQL. 
As illustrated in Figure~\ref{tab:bird_ablation}, when we use Full Schema, the performance of \method on BIRD decreases and results in more token consumption. Therefore, we conclude that redundant schema information in few-shot examples is unnecessary.
Similarly, we conduct ablation studies on the foreign key and cell value information of the test questions. Given that CSM emphasizes schema information, we hypothesize that foreign keys and column values that are not emphasized in the tables could be redundant. 
Experimental results indicate that including all cell values and foreign keys does indeed lead to performance degradation. Notably, including full foreign keys causes a performance drop of nearly 1\% and also increases token consumption.

\begin{figure}[htbp]
    \centering
    \includegraphics[width=0.9\linewidth]{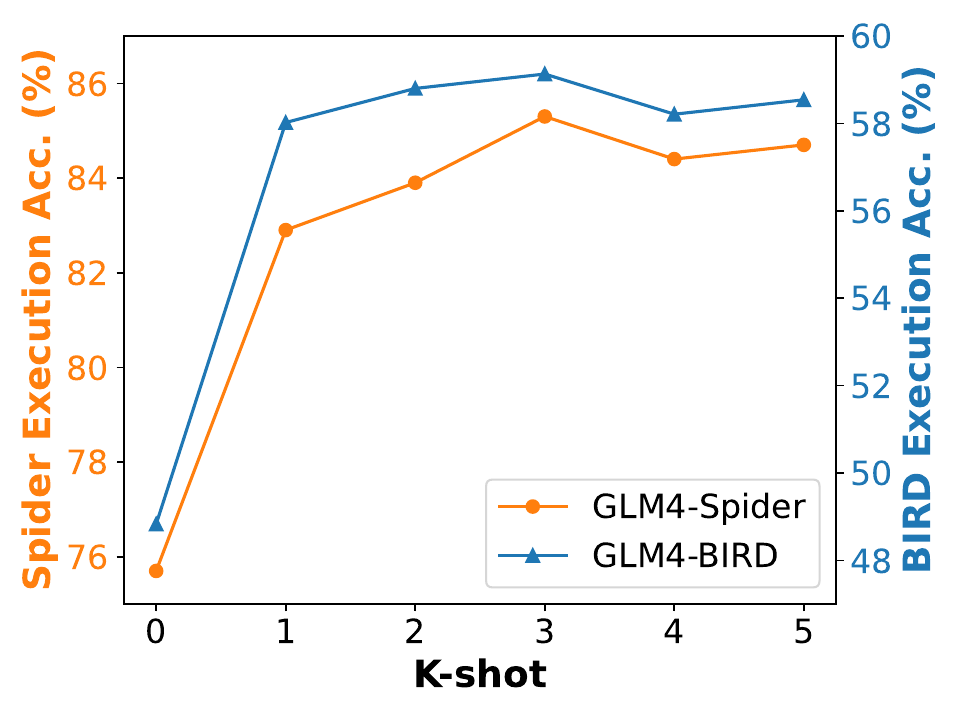}
    \caption{GLM-4' results on various few-shot numbers.}
    \label{fig:fewshot_number}
    \vspace{-4mm}
\end{figure}

\begin{figure*}[htbp]
    \centering
    \includegraphics[width=0.9\linewidth]{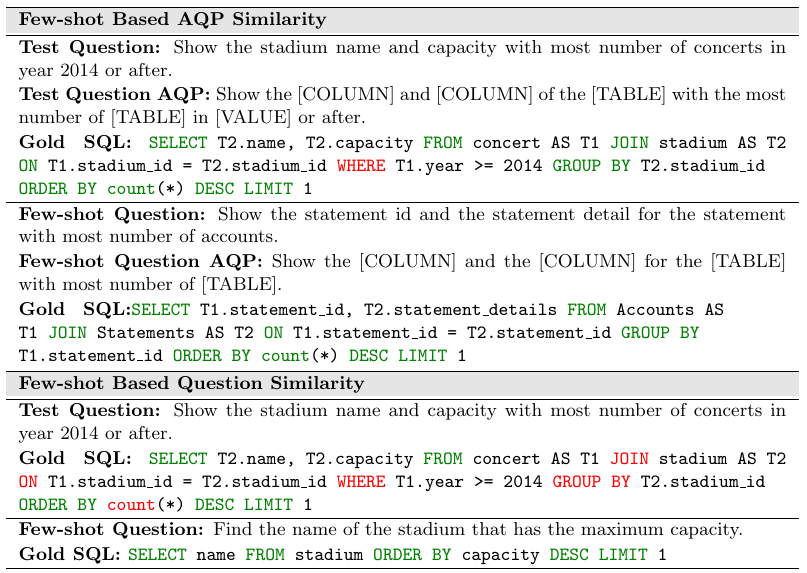}
    \caption{Results of few-shot example retrieval.}
    \label{fig:aqp_case}
    \vspace{-4mm}
\end{figure*}

\subsection{Further Analysis}

\subsubsection{Impacts of Few-shot Numbers}

In the in-context learning of large models, using too few examples may not provide enough reference information for generating accurate SQL queries. Conversely, using multiple examples can increase computational complexity and introduce noise, ultimately reducing performance.
As shown in Figure~\ref{fig:fewshot_number}, we conduct experiments with different numbers of few-shot examples on both BIRD and Spider datasets. 
EX performance improves as the number of few-shot examples increases from 0 to 3 across both datasets. 
However, as the number of examples is larger than 3, performance begins to decline. Thus, three examples are the best choice.

\subsubsection{Case Study}

Our method, by abstracting the samples, can find more valuable few-shot examples for test questions. 
We conduct a detailed case study to illustrate this. 
Figure~\ref{fig:aqp_case} illustrates the top-1 few-shot examples retrieved based on question similarity. The result indicates that \textbf{the SQL structure of the few-shot examples significantly differs from the gold SQL of the test question. }
This discrepancy arises because the examples retrieved based on question similarity often contain abundant domain-specific information, making the retrieval results more domain-focused rather than addressing the specific intent of the question. 
In addition, even if the examples and the test question belong to the same domain, the examples provide little help for SQL generation for the test question due to different databases.

In contrast, \textbf{the top-1 few-shot example retrieved using our proposed AQP shows a high similarity in SQL structure with the test question}, differing only in the ``WHERE'' clause. 
This is because AQP representations provides a database-agnostic representation of the questions, focusing more on the structure of the questions. 
We claim that similar question structures will lead to similar SQL structures. 
Therefore, compared to retrieval based on question similarity, examples obtained through AQP-based retrieval are more helpful in generating the correct SQL for test questions. Ablation results of AQP in Table~\ref{tab:bird_different_level} aslo demonstrates the critical role of AQP.

\subsubsection{CoT version of \method}

\method requires calling the LLM multiple times, which is costly. Therefore, we propose a Chain-of-Thought (CoT) version of \method, which generates AQP, CSM, and initial SQL in a single LLM call.
We evaluate the CoT of \method on the Spider Dev set.
As shown in Table~\ref{tab:cot}, the CoT version's execution accuracy (EX) is only 2.5 percentage points lower than \method+ GPT-4o, but it reduces the average token usage per question by 1,310 tokens. We introduced an efficiency rate to quantify the ratio of model performance to economic cost. 
The results show that the CoT version is more economical. Its efficiency rate is significantly higher than \method, reflecting better optimization in terms of computational resource consumption and performance.

\begin{table}[htbp] 
\centering 
\resizebox{\columnwidth}{!}{ 
\begin{tabular}{lcccc} 
\toprule \textbf{Method} & \textbf{Avg. Prompt Tokens} & \textbf{EX} & \textbf{Efficiency Rate} \\ \midrule \method + GPT-4o & 3,614 & 87.9 & 0.0243 \\ CoT of \method + GPT-4o & 2,304 & 85.4 & 0.0370 \\ \bottomrule 
\end{tabular}
}
    \caption{Performance comparison of different models on execution accuracy and efficiency. Efficiency Rate is calculated by EX/Avg. Prompt Tokens)}
    \label{tab:cot}
\end{table}

%% file: tables/bird_dev.tex
\begin{table}[htbp] 
\small 
\centering 
\resizebox{0.9\columnwidth}{!}{ 
\begin{tabular}{lcc}

\toprule \multirow{2}{*}{Method} & \multicolumn{2}{c}{\textbf{Dev}} \\ \cmidrule(lr){2-3} & \textbf{EX} & \textbf{VES} \\ \midrule 
Palm-2 & 27.38 & - \\ 
ChatGPT + CoT & 36.64 & 42.30 \\ 
Claude-2 & 42.70 & - \\ 
DIN-SQL + GPT-4 & 50.72 & 58.79 \\ 
DAIL-SQL + GPT-4 & 54.76 & 56.08 \\ 
MAC-SQL + GPT-4 & 57.56 & 57.60 \\ 
MAC-SQL + GPT-4o & 57.11 & 60.38 \\
TA-SQL + GPT-4 & 56.19 & - \\ 
SuperSQL & 58.50 & \underline{61.99} \\ \midrule 
GLM-4 & 48.83 & 49.11 \\ 
\textbf{\method + GLM-4} & \underline{59.13} & 59.33 \\ 
GPT-4o & 51.30 & 52.37 \\ 
\textbf{\method + GPT-4o} & \textbf{64.67} & \textbf{65.04} \\ 
\bottomrule 
\end{tabular} 
}
  \caption{Evaluation results on BIRD Dev dataset.}
  \label{tab:bird_dev_results}
  \vspace{-4mm}
\end{table}

%% file: tables/spider_results.tex
\begin{table}[h] 
\small 
\centering 
\resizebox{0.95\columnwidth}{!}{ 
\begin{tabular}{lccc} 
\toprule \textbf{Model} & \textbf{Dev} & \textbf{Test} \\ \midrule ChatGPT-SQL + ChatGPT  & 72.3 & - \\ 
RATSQL + GAP + NatSQL & 75.0 & 73.3 \\ 
T5-3B + PICARD  & 79.3 & 75.1 \\ 
Graphix-3B + PICARD  & 81.0 & 77.6 \\ 
SC-Prompt + T5-3B  & 81.1 & - \\ 
REDSQL-3B + NatSQL & 84.1 & 79.9 \\ 
DIN-SQL + GPT-4  & 82.8 & 85.3 \\ 
MAC-SQL + GPT-4 & 81.4 & 82.8 \\ 
MAC-SQL + GPT-4o & 82.0 & 82.4 \\ 
DAIL-SQL + GPT-4 & 84.4 & \underline{86.6} \\ \midrule 
GLM-4 & 75.7 & - \\ 
\textbf{\method + GLM-4} & \underline{85.3} & 84.3 \\ 
GPT-4o & 75.0 & - \\ 
\textbf{\method + GPT-4o} & \textbf{87.9} & \textbf{86.8} \\ 
\bottomrule 
\end{tabular}
}
\caption{Evaluation results on Spider dataset.}
\label{tab:spider_results}
\vspace{-4mm}
\end{table}

%% file: tables/bird_different_level.tex
\begin{table}[h] 
\centering 
\resizebox{1\columnwidth}{!}{ \begin{tabular}{lccc>{\bfseries}c} \toprule \textbf{Method} & \textbf{Simple} & \textbf{Moderate} & \textbf{Challenging} & \textbf{Total} \\ \midrule
\textbf{\method} & 66.27 & 48.49 & 47.59 & 59.13 \\ 
\ \ \textbf{-w/o CSM} & 62.81 & 46.55 & 44.83 & 56.19 (-2.94) \\ 
\quad \textbf{-w/o AQP} & 61.95 & 43.97 & 41.38 & 54.56 (-4.57) \\ \bottomrule 
\end{tabular}
}
\caption{Ablation results of main components on BIRD.}
\label{tab:bird_different_level}
\vspace{-4mm}
\end{table}

%% file: tables/bird_ablation.tex
\begin{table}[htbp] 
\centering  
\resizebox{0.8\columnwidth}{!}{ 
\begin{tabular}{lc} \toprule \textbf{Method} & \textbf{EX} \\ \midrule \method & 59.13 \\ \midrule w/ Few-shot Full Schema & 58.67 ($\downarrow$) \\ w/ Full Value & 58.41 ($\downarrow$) \\ w/ Full Foreign Key & 58.28 ($\downarrow$) \\ 
\bottomrule 
\end{tabular}
}
\caption{Ablation results of minor components on BIRD.}
\label{tab:bird_ablation}
\vspace{-4mm}
\end{table}

%% file: 2_related.tex
\section{Related Work}

Text-to-SQL aims to convert users' natural language queries into appropriate SQL queries, enabling non-experts to access databases and retrieve information with reduced effort and cost. Early research mainly applies rule-based methods and template matching techniques to translate natural language questions into SQL queries~\citep{baik2020duoquest,stone2020athena++}. However, these methods lack scalability and adaptability.

\vpara{Sequence-to-Sequence based methods.} With the advent of deep learning, Text-to-SQL has evolved into utilizing sequence-to-sequence models~\citep{sutskever2014sequence} where both the database schema and user questions are encoded as sequences, aiming to generate the corresponding SQL query. Transformer-based models, such as T5 and BERT, have enhanced their performance by incorporating relation-aware self-attention mechanisms.

Recently, the advent of LLMs has led to significant advancements in Text-to-SQL task~\citep{rajkumar2022evaluating,ni2023lever,gao2024text}. 
Methods based on LLMs can be roughly classified into \textbf{fine-tuning based methods} and \textbf{prompt-based methods}.
Fine-tuning based methods mainly involve further training open-source language models using Text-to-SQL data~\citep{li2024codes,pourreza2024dts}. 
Prompt-based methods utilize the in-context learning ability of closed-source language models to accomplish text-to-SQL tasks~\citep{pourreza2024din,gao2024text,li2024dawn,qu2024before}.
Fine-tuning-based methods usually require a certain amount of labeled data and computational resources.

\vpara{Prompt-based methods.}
Some early works explore strategies for effectively representing databases in in-context learning
~\cite{rajkumar2022evaluating,chang2023prompt,tai2023exploring,tai2023exploring}. 
In addition, DAIL-SQL~\citep{gao2024text} conducts a comprehensive and systematic evaluation of prompt-based methods, including different forms of information organization and various few-shot retrieval methods.
Subsequent work breaks down this task into multiple stages for solving.
DIN-SQL~\citep{pourreza2024din} breaks down the Text-to-SQL task into smaller subtasks with specific prompts, then guides GPT-4 to complete each subtask, and eventually form a complete SQL query. 
TA-SQL~\citep{qu2024before} introduces Task Alignment, a strategy that enhances large language models' performance and reliability in text-to-SQL tasks by mitigating hallucination through schema linking and logical synthesis.
SuperSQL~\citep{li2024dawn} presents NL2SQL360, a multi-faceted framework for evaluating and comparing natural language to SQL methods to help researchers and practitioners identify optimal solutions for specific needs. 

However, these studies often focus on prompt organization or task decomposition and do not consider the gap between few-shot examples and test questions, which leads to inefficient in-context learning. 
To mitigate these challenges, we propose \model, a SQL generation workflow tailored for real-world and complex database environments.

%% file: 5_conclusion.tex
\section{Conclusion}

In this paper, we analyze and identify two important gaps between questions and SQL queries: the structural mapping gap and the lexical mapping gap. 
We propose \model, an efficient SQL generation pipeline based on LLMs, which alleviates two gaps through Abstract Query Pattern and Contextual Schema Markup. 
Our method achieves leading execution accuracy on the Spider and BIRD datasets. 
Our findings highlight the importance of training corpora. 
We hope that these insights will provide valuable guidance for further research and practical applications in the Text-to-SQL field, and will help to advance its development.

\section{Limitations}
In our work, we do not decompose test questions into sub-questions. When test questions are overly complex, such as those with multiple nested sub-questions, generating the correct SQL becomes more challenging. Additionally, although the training sets of BIRD and Spider are significantly larger than the test sets, this does not guarantee that the training set's AQP can cover all the AQP of the test set questions. If the training set lacks data that matches the AQP of the current test question, performance is adversely affected.

%% file: 6_appendix.tex
\onecolumn

\section{Appendix}
\label{sec:appendix}

Below are prompt templates used for different modules of \model. The last template references the work of MAC-SQL~\citep{wang2024mac}.

\begin{myverbatim}{Prompt Template of Abstract Query Pattern}
You are now an excellent SQL writer. First, I will provide an instruction and three examples. Your task is to learn from the examples how to transform the Original Question into the Masked Question using the DB schema. After that, you will receive a new question with a similar Original Question structure to the examples. Your goal is to replicate the examples to generate the final correct Masked Question.

Let's work this out step by step to ensure we have the right answer. Given a DB schema and an Original Question, follow these steps:

1. Based on the provided schema, identify the required tables and columns in the Original Question for masking.
2. Mask the Original Question with placeholders:
    - Replace table names with [TABLE].
    - Replace column names with [COLUMN].
    - Replace specific values with [VALUE].

Here are three examples:

{ex}

Refer to the examples and respond with the Masked Question with no explanation.

{schema}

###Foreign keys
{db_fk}

### Original Question: {question}
### Masked Question:
\end{myverbatim}

\begin{myverbatim}{Prompt Template of Contextual Schema Markup}
You are now an excellent SQL writer. First, I will provide an instruction and three examples. Your task is to learn from the examples how to transform the Masked Question into the Replaced Question using the DB schema. After that, you will receive a new question with a similar Masked Question structure to the examples. Your goal is to replicate the examples to generate the final correct Replaced Question.

Let's work this out step by step to ensure we have the right answer. Given a DB schema and an Original Question, along with a Masked Question, follow these steps:

1. Identify and replace masked parts such as [COLUMN], [TABLE], and [VALUE] with the appropriate names and values from the schema.
2. For each masked part:
    - If it is a [COLUMN], replace it with the format (masked part from the Original Question, [table].[column]).
    - If it is a [TABLE], replace it with the format (masked part from the Original Question, [table]).
    - If it is a [VALUE], replace it with the format (masked part from the Original Question, [value]).
3. Append additional table and column information that might not have been explicitly mentioned in the original Masked Question but are needed when generating SQL, selecting up to 10 relevant pieces of information.

Here are three examples:

{ex}

Refer to the examples and respond with the Replaced Question with no explanation.

### Schema:
{schema}

### Foreign keys:
{db_fk}

### Original Question: {question}
### Masked Question: {mask}
### Replaced Question:
\end{myverbatim}

\begin{myverbatim}{Prompt Template of Generating SQL}
You are now an excellent SQL writer. First, I will provide an instruction and three examples. Your task is to learn from the examples how to transform the Replaced Question into the Gold SQL using the DB schema. After that, you will receive a new question with a similar Masked Question structure to the examples. Your goal is to replicate the examples to generate the final correct Gold SQL.

Let's work this out step by step to ensure we have the right answer. Given a DB schema and an Original Question, along with a Masked Question and a Replaced Question, follow these steps:

1. Understand the Masked Question:
    - The Masked Question is a version of the Original Question where table names, column names, and specific values are replaced with placeholders such as [TABLE], [COLUMN], and [VALUE].
    - This helps to abstract the question so that it can be mapped to the schema more easily.

2. Understand the Replaced Question:
    - The Replaced Question is derived from the Masked Question by replacing the placeholders with actual table names, column names, and values from the DB schema.
    - Each placeholder is replaced with the format (masked part from the Original Question, [table].[column] for columns, [table] for tables, and [value] for values).
    - At the end of the Replaced Question, additional tables and columns that were not explicitly mentioned in the original question will be appended in the format "Other tables and columns:".

3. Generate the Gold SQL:
    - Identify the necessary tables and columns involved in the Replaced Question.
    - Based on the Replaced Question, analyze how the Gold SQL is constructed.
    - Ensure that the SQL query accurately reflects the intent of the Original Question.

Here are three examples:

{ex}

Refer to the examples and respond with the Gold SQL with no explanation.

{schema}

###Foreign keys
{db_fk}

### Table Value
{value}

### Original Question: {question}
### Masked Question: {mask}
### Replaced Question: {replace}
Pay special attention to the information within the parentheses () in the Replaced Question, as it is crucial for generating the correct Gold SQL.
### Gold SQL:
\end{myverbatim}

\clearpage
\begin{myverbatim}{Prompt Template of Correcting SQL}
[Instruction]
When executing SQL below, some errors occurred, please fix up SQL based on query and database info.
Solve the task step by step if you need to. Using SQL format in the code block, and indicate script type in the code block.
When you find an answer, verify the answer carefully. Include verifiable evidence in your response if possible.
[Constraints]
- In `SELECT <column>`, just select needed columns in the [Question] without any unnecessary column or value
- In `FROM <table>` or `JOIN <table>`, do not include unnecessary table
- If use max or min func, `JOIN <table>` FIRST, THEN use `SELECT MAX(<column>)` or `SELECT MIN(<column>)`
- If [Value examples] of <column> has 'None' or None, use `JOIN <table>` or `WHERE <column> is NOT NULL` is better
- If use `ORDER BY <column> ASC|DESC`, add `GROUP BY <column>` before to select distinct values
[Query]
-- {query}
[Evidence]
{evidence}
[Database info]
{desc_str}
[Foreign keys]
{fk_str}
[old SQL]
```sql
{sql}
```
[SQLite error] 
{sqlite_error}
[Exception class]
{exception_class}

Now please fixup old SQL and generate new SQL again.
[correct SQL]
\end{myverbatim}